\documentclass{article}
\PassOptionsToPackage{numbers,sort&compress}{natbib}
\usepackage[preprint]{neurips_2026}

\usepackage[utf8]{inputenc}
\usepackage[T1]{fontenc}
\usepackage[hidelinks]{hyperref}
\usepackage{url}
\usepackage{booktabs,multirow,makecell,array,tabularx}
\usepackage{amsmath,amsfonts,amssymb}
\usepackage{nicefrac}
\usepackage{microtype}
\usepackage[table]{xcolor}
\usepackage{graphicx}
\usepackage{placeins}
\usepackage{pifont}
\usepackage{fontawesome5}
\usepackage{accsupp}
\definecolor{TableHeader}{HTML}{F0EDF5}
\definecolor{TableBest}{HTML}{CFE2FF}
\definecolor{TableSecond}{HTML}{E8F1FF}
\definecolor{TableSection}{gray}{0.94}
\newcolumntype{L}{>{\raggedright\arraybackslash}X}
\newcolumntype{R}{>{\raggedleft\arraybackslash}X}
\newcolumntype{C}{>{\centering\arraybackslash}X}
\newcommand{\TableSetup}[1]{%
  \rmfamily\upshape\fontsize{#1}{11.4}\selectfont
  \color{black}\arrayrulecolor{black}%
  \setlength{\tabcolsep}{2.3pt}%
  \renewcommand{\arraystretch}{1.12}%
  \setlength{\extrarowheight}{0pt}%
  \setlength{\heavyrulewidth}{0.8pt}\setlength{\lightrulewidth}{0.5pt}%
  \setlength{\cmidrulewidth}{0.3pt}%
  \setlength{\aboverulesep}{0.4ex}\setlength{\belowrulesep}{0.65ex}%
}
\newcommand{\TableHead}[1]{\textbf{#1}}
\newcommand{\TableBestCell}[1]{\cellcolor{TableBest}\textbf{#1}}
\newcommand{\TableRunnerUp}[1]{\cellcolor{TableSecond}\underline{#1}}

\newcommand{\TableGroup}[1]{\textit{#1}}
\newcommand{\TableUnit}[1]{{\fontsize{8}{9}\selectfont #1}}
\newcommand{\TablePanel}[2]{\textbf{#1}\enspace\textit{#2}}

\definecolor{TableHeaderTop}{HTML}{F1F2F3}
\definecolor{TableHeaderSub}{HTML}{F7F8F9}
\definecolor{TableSoftBlue}{HTML}{F0F6FF}
\definecolor{TableSoftSecond}{HTML}{F8FBFF}
\definecolor{TableCoralBest}{HTML}{FCEFEB}
\definecolor{TableCoralSecond}{HTML}{FEF7F5}
\newcommand{\TableRefineSetup}[1]{%
  \TableSetup{#1}%
  \colorlet{TableBest}{TableSoftBlue}\colorlet{TableSecond}{TableSoftSecond}%
  \setlength{\aboverulesep}{0pt}\setlength{\belowrulesep}{0pt}%
  \renewcommand{\arraystretch}{1.2}%
}
\makeatletter
\newcommand{\TableDividerSpan}[1]{\hfill\raisebox{0pt}[0pt][0pt]{%
  \rule[\dimexpr2pt-\dp\@arstrutbox\relax]{.35pt}%
    {\dimexpr#1\ht\@arstrutbox+#1\dp\@arstrutbox-4pt\relax}}}
\makeatother
\newcommand{\TableGroupTag}[1]{\underline{\textit{#1}}}
\newcommand{\TableMetric}[1]{\mbox{\fontsize{8.5}{11.4}\selectfont #1}}
\definecolor{TableGainCoral}{RGB}{218,66,57}
\definecolor{TableGainStrong}{RGB}{184,47,40}
\newcommand{\TableGainInk}{TableGainCoral}
\newcommand{\TableScoreGain}[2]{\mbox{%
  \makebox[18pt][r]{#1}\kern.8pt%
  \raisebox{-.65ex}{\makebox[30pt][l]{\fontsize{6.8}{7.4}\selectfont%
    \if\relax\detokenize{#2}\relax\else%
      \textcolor{\TableGainInk}{\textuparrow\,#2}\fi}}}}
\newcommand{\TableDetectionStrut}{\rule[-5.8pt]{0pt}{16pt}}

\newcommand{\datasetname}{R2V360}
\newcommand{\dtgs}{D$^2$GS}

\newcommand{\AuthorSymbol}[1]{%
  \ifcase#1\relax
  \or\ensuremath{\ast}
  \or\textdagger
  \or\textdaggerdbl
  \or\textsection
  \or\textparagraph
  \or\ensuremath{\star}
  \or\ensuremath{\diamond}
  \or\ding{41}
  \or\ensuremath{\circ}
  \fi}
\DeclareRobustCommand{\CorrespondingSymbol}{%
  \BeginAccSupp{method=hex,unicode,ActualText=2709}%
  \resizebox{!}{.68em}{\faEnvelope[regular]}%
  \EndAccSupp{}}

\newcommand{\AffYichen}{1,2,3}
\newcommand{\AffZihan}{1,2,4}
\newcommand{\AffHao}{1,2,3}
\newcommand{\AffRunyi}{1,2,3}
\newcommand{\AffRuixue}{1,2}
\newcommand{\AffJing}{1,2}
\newcommand{\AffYunong}{1,2}
\newcommand{\AffLei}{1,2}
\newcommand{\AffKeqiang}{1,2}
\newcommand{\AuthorName}[2]{\mbox{\textbf{#1}\textsuperscript{\normalfont #2}}}
\newcommand{\AuthorRowsWide}{%
  \begingroup\fontsize{10}{13}\selectfont
  \AuthorName{Yichen Zhang}{\AffYichen}\hspace{.9em}%
  \AuthorName{Zihan Wang}{\AffZihan}\hspace{.9em}%
  \AuthorName{Hao Chen}{\AffHao}\hspace{.9em}%
  \AuthorName{Runyi Huang}{\AffRunyi}\hspace{.9em}%
  \AuthorName{Ruixue Zong}{\AffRuixue}\par
  \vspace{4pt}
  \AuthorName{Jing Wang}{\AffJing}\hspace{1.4em}%
  \AuthorName{Yunong Mao}{\AffYunong}\hspace{1.4em}%
  \AuthorName{Lei He}{\AffLei,\CorrespondingSymbol}\hspace{1.4em}%
  \AuthorName{Keqiang Li}{\AffKeqiang}\par\endgroup}
\newcommand{\Affiliations}{%
  {\fontsize{9}{11}\selectfont
  \textsuperscript{1}The School of Vehicle and Mobility, Tsinghua University\par
  \mbox{\textsuperscript{2}The State Key Laboratory of Intelligent Green Vehicle and Mobility, Tsinghua University}\par
  \textsuperscript{3}College of AI, Tsinghua University\par
  \textsuperscript{4}Shanghai Jiao Tong University\par}}
\newcommand{\GroupedEmails}{%
  {\normalfont\fontsize{8}{10}\selectfont\ttfamily\setlength{\parskip}{0pt}%
  \{yichen-zhang26,chenhao26,hry21\}@mails.tsinghua.edu.cn\par
  \{rx-zong,helei2023,likq\}@tsinghua.edu.cn\par}}
\newcommand{\AuthorBlock}{%
  \begingroup\normalfont\fontsize{11}{14}\selectfont\setlength{\parskip}{0pt}%
  \AuthorRowsWide
  \vspace{8pt}\Affiliations
  \vspace{6pt}\GroupedEmails
  \endgroup}
\author{\AuthorBlock}
\makeatletter
\renewcommand{\@maketitle}{%
  \vbox{\hsize\textwidth\linewidth\hsize
    \vskip 0.1in\@toptitlebar\centering
    {\LARGE\bfseries\@title\par}\@bottomtitlebar
    \vskip 12pt\@author
    \vskip 0.3in\@minus 0.1in}}
\makeatother
\newcommand{\CorrespondingAuthorNote}{%
  \begingroup
  \renewcommand{\thefootnote}{\CorrespondingSymbol}%
  \footnotetext[1]{Corresponding author.}%
  \endgroup}

\title{Beyond Vehicle-Centric Data: Scaling Autonomous Driving through Roadside-to-Vehicle Generation}
\hypersetup{pdftitle={Transforming roadside observations into vehicle training data for autonomous driving},
  pdfauthor={Yichen Zhang, Zihan Wang, Hao Chen, Runyi Huang, Ruixue Zong, Jing Wang, Yunong Mao, Lei He, Keqiang Li},
  pdfsubject={Preprint}}

\begin{document}
\maketitle
\CorrespondingAuthorNote

\begin{abstract}
Training end-to-end autonomous driving models requires large, diverse real-world driving datasets, collected primarily by vehicles. Scaling these datasets entails sustained investment in fleet operation, data processing and annotation. Road infrastructure also accumulates real traffic records that capture the motion and interactions of multiple vehicles, yet the substantial viewpoint gap between roadside and onboard sensors limits their direct use for training. Here we introduce AnyR2V, a framework that transforms multimodal roadside observations into continuous seven-camera surround-view videos from different vehicle perspectives. The framework aggregates sequential roadside observations into a shared scene representation that captures geometry and road-user motion, then renders appearance, depth and layout conditions along target-vehicle trajectories. With supervision from spatiotemporally aligned roadside--vehicle pairs, AnyR2V learns to jointly generate surround-view videos from these incomplete conditions, guided by recorded scene structure and road-user motion. Existing roadside three-dimensional (3D) bounding boxes are transformed into target-vehicle coordinates and paired with the generated images, allowing a single traffic recording and its annotations to yield training data from multiple vehicle perspectives. To support training and evaluation, we construct R2V360, a dataset of 89 real traffic sequences recorded at an urban intersection, with spatially and temporally aligned multimodal roadside data and surround-view videos from a recording vehicle. AnyR2V outperforms six scene reconstruction methods in image fidelity and in the agreement of depth and semantic predictions between generated and real onboard images. We further assess the training value of surround-view videos generated along other vehicles' trajectories by adding them to real training data. Adding generated frames equivalent to 50\% of the original real training frame count improves mean average precision (mAP) by 3.3--3.4 AP-points across three 3D object detectors on the real onboard test set, relative to real-data-only baselines. These results demonstrate that traffic recorded by road infrastructure can provide effective training data from multiple vehicle perspectives, offering a new data source for autonomous driving.

\end{abstract}

\section{Introduction}
\label{sec:anyr2v-introduction}

Road traffic crashes remain a major global public safety concern, claiming approximately 1.19 million lives each year according to the World Health Organization~\cite{WHORoadSafety2023}. Autonomous driving offers the prospect of reducing human driving errors and improving road safety, provided that systems operate safely and reliably amid the complexity and variability of real traffic. End-to-end autonomous driving has emerged as a prominent approach, integrating perception, prediction and planning within a unified neural network trained on large-scale real-world driving data~\cite{PlanningOrientedAutonomousDriving,EndToEndAutonomousDriving}. In this paradigm, the ability to handle complex traffic depends strongly on the scale, quality and coverage of real-world onboard sensor data~\cite{EndToEndAutonomousDriving}. Commercial deployments illustrate the scale involved: Tesla reported that, by March 2026, its fleet had accumulated nearly 10 billion miles of supervised driving~\cite{TeslaQ12026Update}. Acquiring such datasets requires sustained fleet operation, together with substantial financial and human resources for data collection, processing and annotation.

In current vehicle-based data pipelines, real-world recordings undergo selection, annotation and quality assurance before model training and evaluation. Challenging scenarios identified during evaluation then guide further targeted collection, completing the data feedback loop. Fleet operation, equipment maintenance and data transmission require sustained investment, as do data storage, cleaning, de-identification and annotation review~\cite{nuScenes}. Successive iterations repeat these collection and processing stages, adding to dataset development time and cost. Coverage is also constrained by the routes and times at which a fleet operates, making rare yet valuable traffic situations difficult to capture promptly~\cite{LearningNaturalisticDrivingEnvironment}. Individual companies build separate data resources around their own fleets, creating silos that constrain sharing and reuse~\cite{RS2ADLiDAR}. A central challenge for data-driven autonomous driving is therefore to efficiently construct large-scale, high-quality training datasets with broad coverage while reducing dependence on continual vehicle-based collection.

Novel-view synthesis through 3D scene reconstruction offers a route to reusing existing onboard data. Methods based on neural radiance fields (NeRF) and 3D Gaussian splatting (3DGS) model static backgrounds and dynamic road users, reconstructing a recorded driving sequence as a renderable scene. Rendering this scene at new camera poses allows the same recording to yield training data from additional viewpoints~\cite{NeuralSceneGraphs,StreetGaussians}. Optimising these representations, however, relies on geometric and appearance supervision from the original observations. Target viewpoints may deviate substantially from the original recording trajectory, for example by following other vehicles’ trajectories in the same traffic sequence. At these viewpoints, some road regions and object surfaces visible to the target camera may fall outside the original observational coverage. These gaps in coverage constrain reconstruction accuracy and the fidelity of novel-view synthesis~\cite{StreetCrafter}.

Generative driving world models provide a complementary route to data augmentation by diversifying scene content and vehicle motion. These models learn visual distributions and temporal dynamics from driving videos, then synthesise new driving sequences conditioned on initial frames, actions, road layouts or 3D bounding boxes~\cite{GAIA1,DriveDreamer}. Geometry-guided generation further combines observed scenes with generative priors to improve control over novel-view synthesis~\cite{StreetCrafter}. Systematic evaluations nevertheless reveal a gap between visual quality and physical and behavioural realism~\cite{WorldLens}. Autonomous driving systems ultimately share real roads with human drivers, making it important for training data to capture how road users move and respond to one another in specific situations~\cite{LearningNaturalisticDrivingEnvironment}. Extending a recorded traffic sequence to additional viewpoints therefore requires generated images to remain aligned with the recorded scene structure and road-user motion.

Both approaches rely primarily on vehicle-based recordings. Scene reconstruction reuses real traffic within a limited range of viewpoints; world models support broader generalisation, with fidelity to real traffic remaining a challenge. Overcoming these limitations calls for reconsidering the observation sources used to construct vehicle-view training data. Roadside sensors provide a complementary source of continuous observations of real traffic~\cite{V2XSeq}. Mounted several metres above the road, they offer an elevated overview of an entire intersection and continuously record the motion of road users within their sensing range, capturing a shared traffic scene involving multiple vehicles. Fixed deployment enables sustained observation of the same area, allowing data to accumulate throughout sensor operation. In interaction-rich locations, continuous coverage also creates opportunities to capture rare yet valuable traffic situations. Collected through public infrastructure, these data can be shared and reused across vehicles independently of any particular fleet, offering a way to overcome the silos created by fleet-specific collection.

RS2AD-LiDAR converts roadside observations into LiDAR point clouds from target-vehicle viewpoints and demonstrates their value for training 3D object detectors~\cite{RS2ADLiDAR}. Here we investigate how multimodal roadside observations can be transformed into continuous surround-view videos for target vehicles, turning the traffic records accumulated by road infrastructure into training data from different vehicle perspectives. Differences in sensor height, orientation and motion between roadside and onboard observations alter visible surfaces and occlusion patterns, leaving gaps in the geometry and appearance available for the target views. The simultaneous motion of the target vehicle and surrounding road users further requires precise correspondence between camera pose and scene time. When constructing paired training samples, spatial errors displace scene projections relative to the real images, whereas temporal errors misalign moving objects. Video generation must therefore translate roadside observations and their spatiotemporal relationships into target-view conditions, completing missing visual content while maintaining scene correspondence across successive frames and surround-view cameras.

To address these challenges, we introduce AnyR2V, which transforms multimodal roadside observations into continuous seven-camera surround-view videos from different vehicle perspectives. The framework first aggregates multiple frames of roadside observations into a shared scene representation comprising the static background, object geometry and motion trajectories. It then renders appearance, depth and layout conditions at the target camera poses and capture times. Using real videos from the recording vehicle as supervision, we fine-tune the control branch of a pretrained multi-view video model~\cite{WorldSimulationWithVideoFoundationModels} to learn a mapping from these incomplete conditions to surround-view videos. For each additional vehicle, its trajectory defines the camera viewpoints, and the same scene representation and generative model are reused.

We construct R2V360 to provide paired data for training and evaluating this transformation. The dataset comprises 89 real traffic sequences from an urban signalised intersection, with calibration and temporal alignment linking multimodal roadside recordings to seven-camera surround-view videos from the recording vehicle. These paired data enable frame-by-frame comparison of generated videos with synchronously recorded onboard videos at matching camera poses and capture times. Trajectories of other vehicles in the roadside recordings support generation from additional viewpoints. Existing roadside 3D bounding boxes are transformed into each target vehicle's coordinate frame and paired with the generated surround-view images, providing directly usable training samples for 3D object detection.

In paired evaluation from the recording vehicle's viewpoints, AnyR2V outperforms all six scene reconstruction baselines in image fidelity and in the agreement of depth and semantic predictions with those obtained from real onboard images. To assess the value of the generated data for real-world onboard perception, we augment the real training set with surround-view videos generated along other vehicles' trajectories and train 3D object detectors on the combined data. Adding generated frames equivalent to 50\% of the original real training frame count improves mean average precision (mAP) by 3.3--3.4 AP-points across three detectors on the real onboard test set, relative to real-data-only baselines. These results demonstrate that roadside recordings can be transformed into effective training data from multiple vehicle perspectives, providing an additional data source for training onboard perception models.


\section{Results}
\label{sec:anyr2v-results}

\subsection{The R2V360 dataset}
\label{sec:anyr2v-results-dataset}

\begin{figure}[!htbp]
\centering
\includegraphics[width=\textwidth]{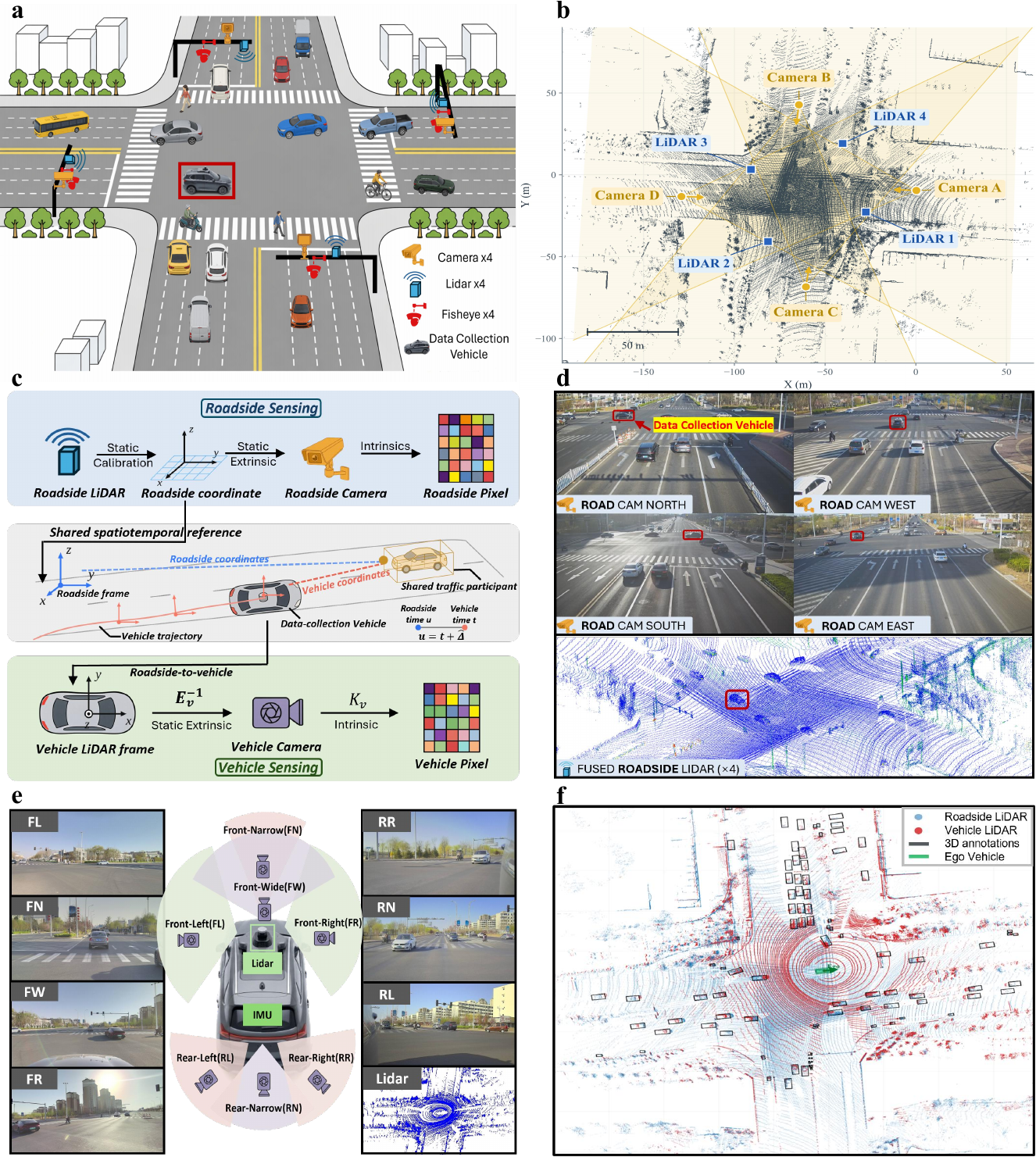}
\caption{\textbf{Roadside and onboard sensor configurations and spatiotemporal data alignment in R2V360.}
\textbf{a} Schematic of roadside sensor deployment and traffic at the urban intersection. The red box identifies the recording vehicle.
\textbf{b} Spatial arrangement of roadside LiDAR sensors and cameras, with camera fields of view. Blue squares and yellow circles mark the LiDAR and camera locations, respectively.
\textbf{c} Spatiotemporal alignment and projection relationships between roadside and onboard data, including sensor calibration, coordinate transformations for the recording vehicle, temporal-offset calibration and projection of 3D points onto camera images.
\textbf{d} Synchronously recorded roadside images from four directions and the fused point cloud from the four roadside LiDAR sensors. Red boxes identify the recording vehicle in the images and point cloud.
\textbf{e} Sensor configuration and example observations from the recording vehicle, which is equipped with seven surround-view cameras, one LiDAR sensor and an inertial measurement unit (IMU). Camera images and the onboard LiDAR point cloud are shown on either side.
\textbf{f} Overlaid roadside and onboard point clouds after spatiotemporal alignment, shown in blue and red, respectively. Black boxes denote 3D object annotations, and the green box identifies the recording vehicle.}
\label{fig:overview}
\end{figure}

We construct R2V360, a paired dataset for roadside-to-vehicle video generation, by synchronously recording multimodal roadside observations and onboard surround-view videos at an urban signalised intersection. The roadside installation comprises four pinhole cameras, four fisheye cameras and four LiDAR sensors distributed around the four approaches to the intersection, and the recording vehicle carries seven cameras and one LiDAR sensor. These configurations provide $360^\circ$ multimodal coverage on both platforms (Fig.~\ref{fig:overview}). Spatial calibration and temporal alignment link the roadside scene to observations from each onboard camera at its capture time. The synchronised onboard surround-view videos provide paired supervision for the generative model and serve as references for evaluating generation quality.

\begin{figure}[!htbp]
\centering
\includegraphics[width=\textwidth]{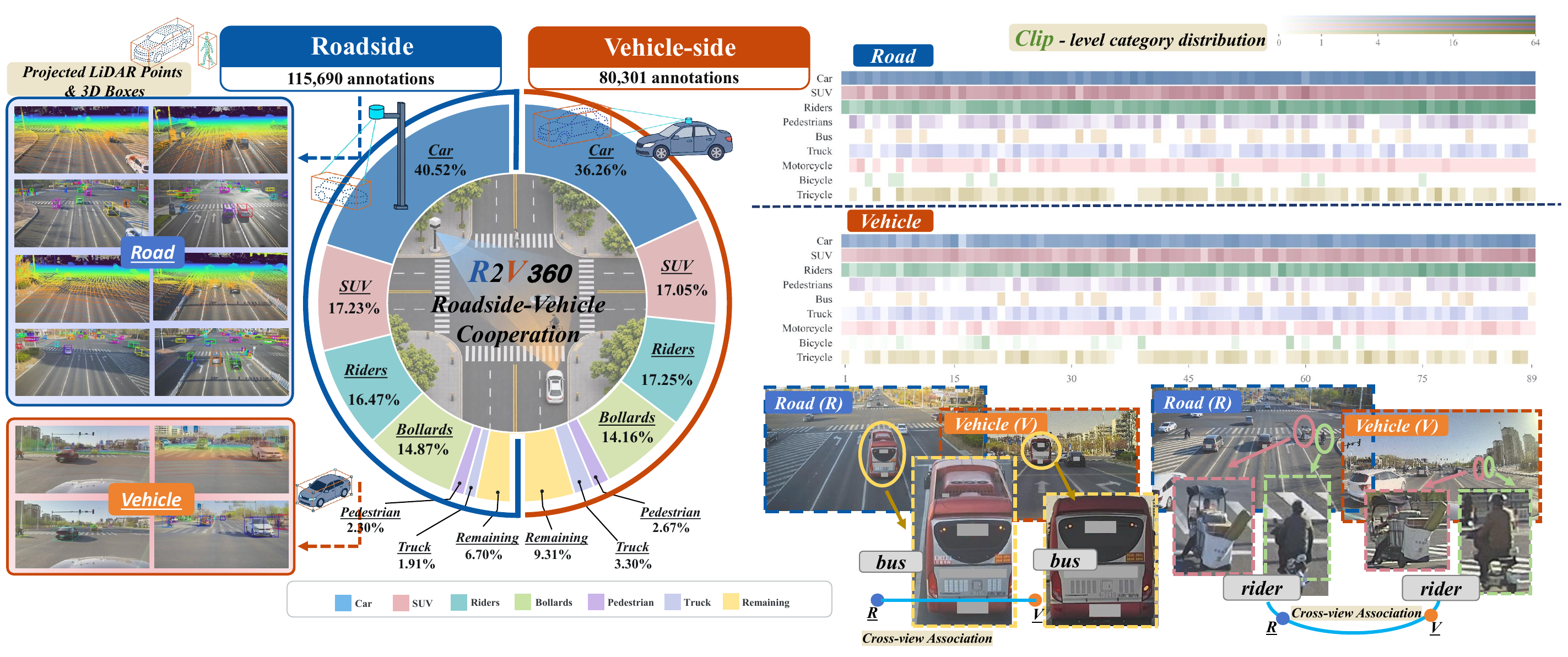}
\caption{\textbf{3D object annotations and category distributions in R2V360.}
Left, projections of LiDAR point clouds and 3D bounding boxes onto synchronously recorded roadside and onboard images.
Centre, doughnut chart showing category proportions among the original keyframe annotations, comprising 115,690 roadside and 80,301 onboard 3D bounding boxes.
Top right, heatmaps of category distributions across the 89 traffic sequences, with roadside and onboard data shown in the upper and lower groups, respectively. Each column represents a traffic sequence and each row an object category; hue distinguishes categories, and intensity indicates the mean number of bounding boxes for each category per keyframe in the corresponding sequence.
Bottom right, corresponding buses and riders in roadside and onboard images, with enlarged views. Circles and connecting lines of the same colour identify the same object.
Blue and orange borders denote roadside and onboard data, respectively.}
\label{fig:dataset-distribution}
\end{figure}

The dataset comprises 89 traffic sequences and 332,215 roadside and onboard images, with 3D annotations spanning 23 categories for 12,891 roadside and 14,972 onboard LiDAR frames. The original keyframes from the two platforms contain a total of 195,991 3D bounding boxes. Their category distribution and roadside--vehicle object correspondences are shown in Fig.~\ref{fig:dataset-distribution}. Including interpolated annotations brings the total to 1,744,624 bounding boxes.

Among the datasets compared in Table~\ref{tab:dataset}, conventional onboard datasets are restricted to vehicle perspectives, and existing real-world vehicle--infrastructure datasets have limited angular coverage on at least one platform. DeepAccident provides $360^\circ$ coverage on both platforms through simulation. R2V360 combines real traffic recordings, surround-view observations from both platforms and object trajectories, providing a paired dataset for research on roadside-to-vehicle video generation.

\begin{table}[!htbp]
\centering
\caption{\textbf{Comparison of R2V360 with existing autonomous driving datasets.} Sensor types and surround-view coverage are listed separately for the roadside and vehicle platforms. C, F and L denote pinhole cameras, fisheye cameras and LiDAR sensors, respectively. V, I, VIC and V2V denote vehicle-only, infrastructure-only, vehicle--infrastructure cooperative and vehicle--vehicle cooperative observations, respectively. A check mark indicates $360^\circ$ coverage by both cameras and LiDAR on the corresponding platform; a cross indicates that at least one modality has more limited coverage. Dashes in the sensor and coverage columns denote the absence of the corresponding platform; dashes in the scale columns indicate values unreported in the original publications. K and M denote thousands and millions, respectively. Light blue highlights R2V360.}
\label{tab:dataset}
\begingroup
\TableRefineSetup{8.5}
\setlength{\tabcolsep}{2.2pt}
\begin{tabularx}{\linewidth}{>{\hsize=2.25\hsize\linewidth=\hsize}L >{\hsize=.9\hsize\linewidth=\hsize}L >{\hsize=1\hsize\linewidth=\hsize}C >{\hsize=1.2\hsize\linewidth=\hsize}C >{\hsize=.55\hsize\linewidth=\hsize}C >{\hsize=.85\hsize\linewidth=\hsize}C >{\hsize=.55\hsize\linewidth=\hsize}C >{\hsize=.8\hsize\linewidth=\hsize}R >{\hsize=.9\hsize\linewidth=\hsize}R}
\toprule
\rowcolor{TableHeaderTop}
 & & & \multicolumn{2}{c}{\cellcolor{TableHeaderTop}\TableHead{Infrastructure}} & \multicolumn{2}{c}{\cellcolor{TableHeaderTop}\TableHead{Vehicle}} & \multicolumn{2}{c}{\cellcolor{TableHeaderTop}\TableHead{Scale}} \\
\rowcolor{TableHeaderSub}
\cellcolor{TableHeaderTop}\multirow{-2}{*}{\TableHead{Dataset}}\TableDividerSpan{2} & \cellcolor{TableHeaderTop}\multirow{-2}{*}{\TableHead{Source}} & \cellcolor{TableHeaderTop}\multirow{-2}{*}{\TableHead{Views}} & Sensors & $360^\circ$ & Sensors & $360^\circ$ & Frames & 3D boxes \\
\midrule
KITTI~\cite{geiger2012kitti} & Real & V & -- & -- & 4C+1L & $\times$ & 15K & 80K \\
nuScenes~\cite{nuScenes} & Real & V & -- & -- & 6C+1L & $\checkmark$ & 200K & 1.4M \\
Waymo Open~\cite{WaymoOpenDataset}\TableDividerSpan{3} & Real & V & -- & -- & 5C+5L & $\checkmark$ & 600K & 12M \\
\midrule
DAIR-V2X-I~\cite{DAIRV2X} & Real & I & 1C+1L & $\times$ & -- & -- & 10K & 493K \\
V2X-Sim~\cite{li2022v2xsim} & Simulated & V2V, VIC & 4C+1L & $\times$ & 6C+1L & $\checkmark$ & 10K & 27K \\
DeepAccident~\cite{wang2023deepaccident} & Simulated & V2V, VIC & 6C+1L & $\checkmark$ & 6C+1L & $\checkmark$ & 57K & -- \\
TUMTraf V2X~\cite{zimmer2024tumtraf} & Real & VIC & 4C+1L & $\times$ & 1C+1L & $\times$ & 2K & 30K \\
DAIR-V2X-C~\cite{DAIRV2X} & Real & VIC & 1C+1L & $\times$ & 2C+2L & $\times$ & 39K & 464K \\
V2X-Seq~\cite{V2XSeq}\TableDividerSpan{6} & Real & VIC & 1C+1L & $\times$ & 2C+2L & $\times$ & 15K & -- \\
\midrule
\rowcolor{TableSoftBlue}
\textbf{\datasetname{} (OURS)}\TableDividerSpan{1} & Real & VIC & 4C+4F+4L & $\checkmark$ & 7C+1L & $\checkmark$ & 28K & 196K \\
\bottomrule
\end{tabularx}
\endgroup
\end{table}

\subsection{Fidelity of the generated videos}
\label{sec:anyr2v-results-generation}

We evaluate the quality of AnyR2V's seven-camera surround-view videos on nine test sequences, using synchronously recorded onboard videos as references and comparing against six scene reconstruction methods based on Gaussian splatting. Each baseline reconstructs the scene from the roadside images and point clouds of the test sequence, then renders images at the same target camera poses and capture times as AnyR2V. The evaluation covers framewise image similarity, overall image distributions, and agreement in depth and semantic predictions. Experimental settings are detailed in Section~\ref{sec:anyr2v-eval-video}.

AnyR2V's main advantages are in perceptual similarity and agreement with the distribution of real onboard images (Table~\ref{tab:gs}). Compared with Co-Adaptation, the strongest baseline on these two metrics, AnyR2V reduces LPIPS by 24.7\% and FID by 58.7\%. Its PSNR also exceeds that of all reconstruction baselines.

\begin{table}[!htbp]
\centering
\caption{\textbf{Fidelity and structural consistency of the generated videos.} All methods are evaluated on the same test image set, using synchronously recorded onboard images as references. Depth and semantic metrics compare predictions from the respective pretrained models on generated and real images. Seg-Cons denotes pixelwise semantic label agreement. Arrows indicate the preferred direction for each metric; bold and underlined values denote the best and second-best results, respectively.}
\label{tab:gs}
\begingroup
\TableRefineSetup{9}
\setlength{\tabcolsep}{1.5pt}
\colorlet{TableBest}{TableCoralBest}\colorlet{TableSecond}{TableCoralSecond}
\begin{tabularx}{\linewidth}{l >{\hsize=1.3\hsize\linewidth=\hsize}C >{\hsize=.85\hsize\linewidth=\hsize}C >{\hsize=.95\hsize\linewidth=\hsize}C >{\hsize=1.4\hsize\linewidth=\hsize}C >{\hsize=.7\hsize\linewidth=\hsize}C >{\hsize=1\hsize\linewidth=\hsize}C >{\hsize=.8\hsize\linewidth=\hsize}C c}
\toprule
 & \multicolumn{3}{c}{\TableGroup{Fidelity}} & \multicolumn{2}{c}{\TableGroup{Realism}} & \multicolumn{2}{c}{\TableGroup{Geometry}} & \TableGroup{Semantics} \\
\cmidrule(lr){2-4}\cmidrule(lr){5-6}\cmidrule(lr){7-8}\cmidrule(lr){9-9}
\rowcolor{TableHeaderSub}
\cellcolor{white}\multirow{-2}{*}{\TableHead{Method}} & \TableMetric{PSNR\,\TableUnit{(dB)}\,$\uparrow$} & \TableMetric{SSIM\,$\uparrow$} & \TableMetric{LPIPS\,$\downarrow$} & \TableMetric{KID\,\TableUnit{($\times10^3$)}\,$\downarrow$} & \TableMetric{FID\,$\downarrow$} & \TableMetric{AbsRel\,$\downarrow$} & \TableMetric{$\delta_1$\,\TableUnit{(\%)}\,$\uparrow$} & \TableMetric{Seg-Cons\,\TableUnit{(\%)}\,$\uparrow$} \\
\midrule
DropGaussian~\cite{wu2024dropgaussian} & 9.89 & 0.382 & 0.913 & 271.63 & 274.4 & 0.639 & 38.6 & 35.4 \\
\dtgs{}~\cite{song2026d2gs} & 9.86 & 0.381 & 0.914 & 261.88 & 271.5 & 0.656 & 38.4 & 35.4 \\
Co-Adaptation~\cite{zhou2025coadaptation} & 10.20 & \TableRunnerUp{0.387} & \TableRunnerUp{0.872} & \TableRunnerUp{229.12} & \TableRunnerUp{238.8} & 0.634 & 38.5 & \TableRunnerUp{38.2} \\
SparseGS~\cite{li2024sparsegs} & 6.61 & 0.179 & 0.951 & 320.35 & 298.9 & 0.575 & 42.8 & 33.2 \\
AD-GS~\cite{hu2024adgs} & \TableRunnerUp{11.51} & 0.378 & 0.887 & 297.10 & 294.3 & \TableRunnerUp{0.537} & \TableRunnerUp{44.5} & 37.6 \\
S3Gaussian~\cite{huang2024s3gaussian} & 6.26 & 0.133 & 0.942 & 318.47 & 309.6 & 0.574 & 39.4 & 31.4 \\
\midrule
\textbf{OURS} & \TableBestCell{14.11} & \TableBestCell{0.389} & \TableBestCell{0.657} & \TableBestCell{38.64} & \TableBestCell{98.7} & \TableBestCell{0.483} & \TableBestCell{54.0} & \TableBestCell{63.9} \\
\bottomrule
\end{tabularx}
\endgroup
\end{table}

We assess depth and semantic agreement using relative depth maps and semantic segmentation maps predicted from generated and real onboard images by pretrained depth estimation and segmentation models. AnyR2V achieves 63.9\% pixelwise semantic label agreement, exceeding the strongest baseline by 25.7 percentage points (Table~\ref{tab:gs}). After scale-and-shift alignment in disparity space, its relative depth error is also lower than that of all baselines. Together, these results show closer agreement with the corresponding predictions from real onboard images than is achieved by the reconstruction baselines.

\begin{figure}[!htbp]
\centering
\includegraphics[width=\textwidth]{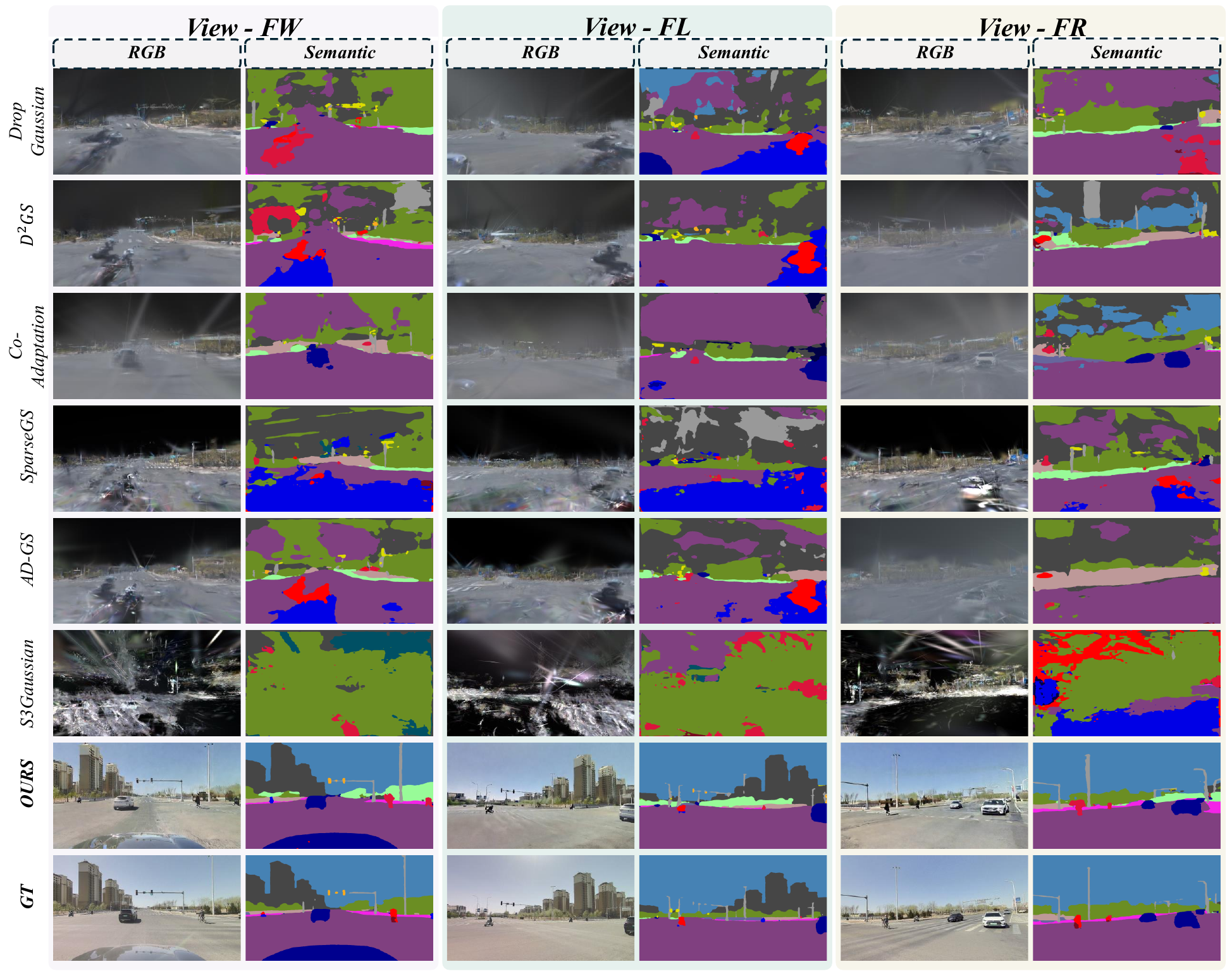}
\caption{\textbf{Qualitative comparison of generated images and their semantic segmentation maps.} Rows show, from top to bottom, the six Gaussian-splatting baselines, AnyR2V (OURS) and real onboard images (GT). The three column groups show RGB images and corresponding semantic segmentation maps for the front wide-angle (FW), front-left (FL) and front-right (FR) views, respectively. All segmentation maps are predicted by the same Mask2Former model.}
\label{fig:qualitative}
\end{figure}

The qualitative comparisons in Fig.~\ref{fig:qualitative} highlight differences in road-surface completeness and the sharpness of moving-vehicle boundaries. Some Gaussian-splatting reconstructions contain fragmented road surfaces or merge vehicles with the background. AnyR2V produces more continuous road surfaces and sharper vehicle contours, with semantic segmentation maps that more closely match those predicted from real onboard images.

\subsection{Surround-view video generation for multiple vehicles}
\label{sec:anyr2v-results-other-vehicles}

We further examine surround-view videos generated for different target vehicles to assess reuse of the same roadside recording across vehicle perspectives and correspondence across successive frames and camera views. For target vehicles without synchronously recorded onboard videos, we use the relationships among road users recorded in roadside images, point clouds and 3D annotations as references.

\begin{figure}[!htbp]
\centering
\includegraphics[width=\textwidth]{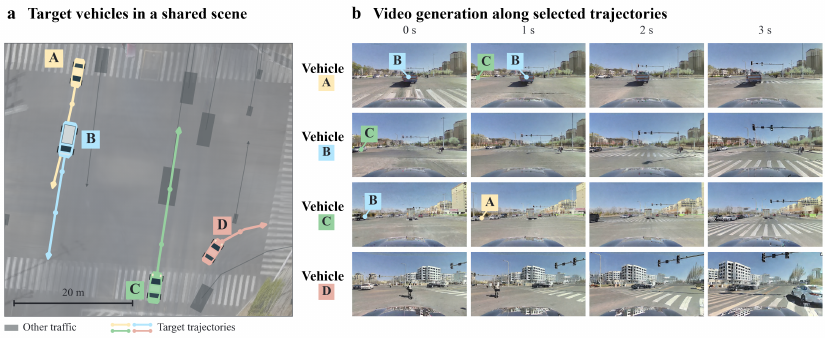}
\caption{\textbf{Multi-vehicle video generation from a single roadside recording.}
\textbf{a} Initial positions and trajectories of target vehicles A--D. The background is a bird's-eye view constructed from roadside images. Coloured markers and trajectories denote target vehicles, and grey indicates other road users. Markers along the trajectories correspond to the times shown in \textbf{b}.
\textbf{b} Generated front wide-angle frames for each target vehicle at 0, 1, 2 and 3\,s. Each row corresponds to one target vehicle, and coloured letters identify the other target vehicles visible in that view. Videos for different vehicles are generated independently.}
\label{fig:anyvehicle}
\end{figure}

The generated views preserve relative vehicle positions observed in the roadside recording (Fig.~\ref{fig:anyvehicle}). Vehicle B is visible ahead of A, and the oncoming vehicle C appears in the views of both A and B; C's own view also shows A and B. As D turns right, its forward view rotates with the vehicle towards the adjoining road. These results show that a single roadside recording can yield traffic videos from different vehicle perspectives while retaining the relative positions of vehicles.

\begin{figure}[!htbp]
\centering
\includegraphics[width=\textwidth]{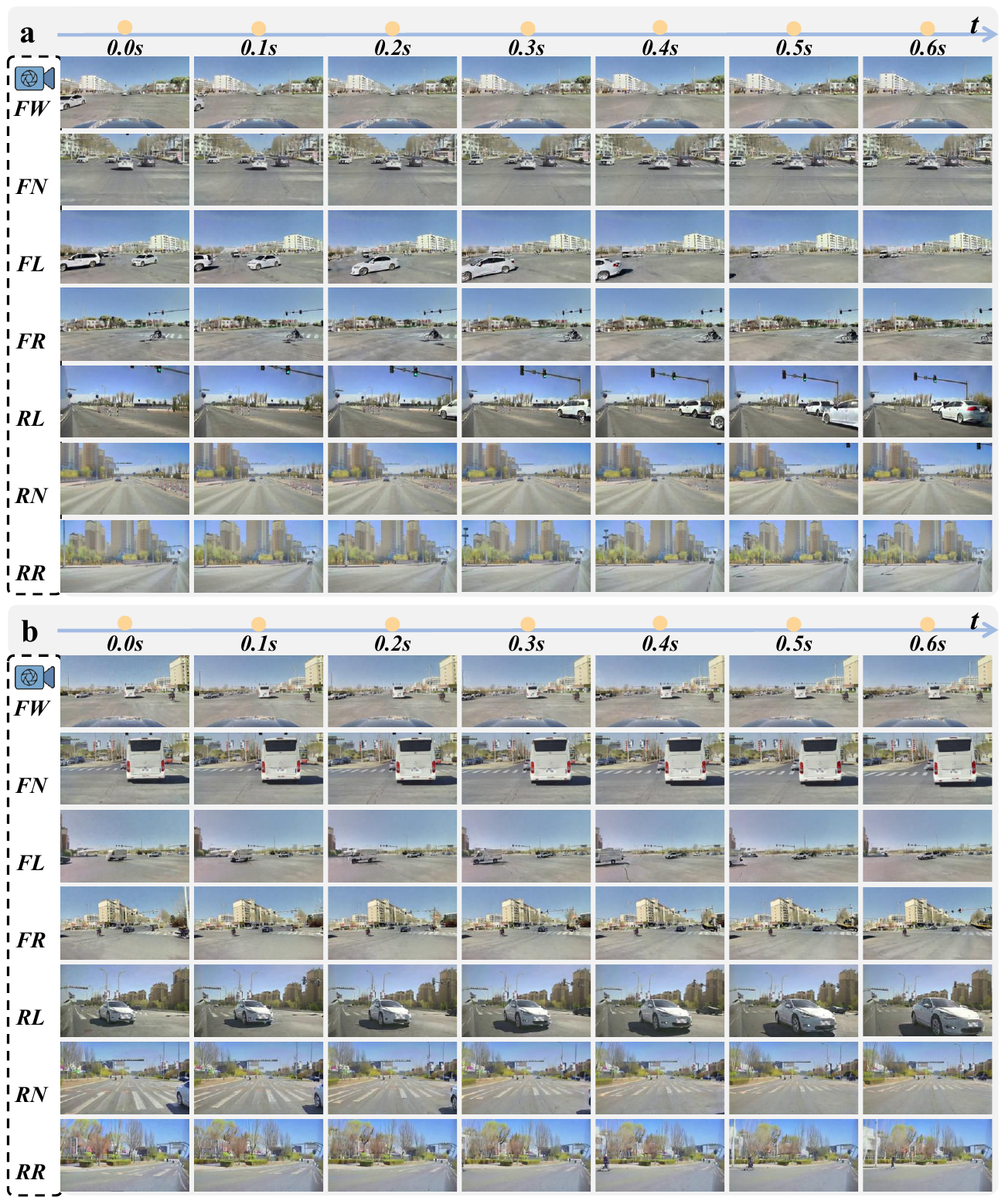}
\caption{\textbf{Temporal continuity and cross-view consistency of seven-camera video generation.}
\textbf{a,b} Two sets of surround-view videos generated from roadside recordings. Rows correspond to camera views, and columns show consecutive sampling times from 0 to 0.6\,s at 0.1\,s intervals. FW, FN, FL, FR, RL, RN and RR denote the front wide-angle, front narrow-angle, front-left, front-right, rear-left, rear narrow-angle and rear-right cameras, respectively.}
\label{fig:sequence}
\end{figure}

Figure~\ref{fig:sequence} shows the temporal evolution of the seven generated camera views. Road structure and vehicle appearance remain coherent across adjacent frames, while the relative positions of surrounding vehicles change continuously with the target vehicle's motion. Views from differently oriented cameras show corresponding road layouts and vehicle positions, together depicting the traffic scene around the target vehicle.

\begin{figure}[!htbp]
\centering
\includegraphics[width=\textwidth]{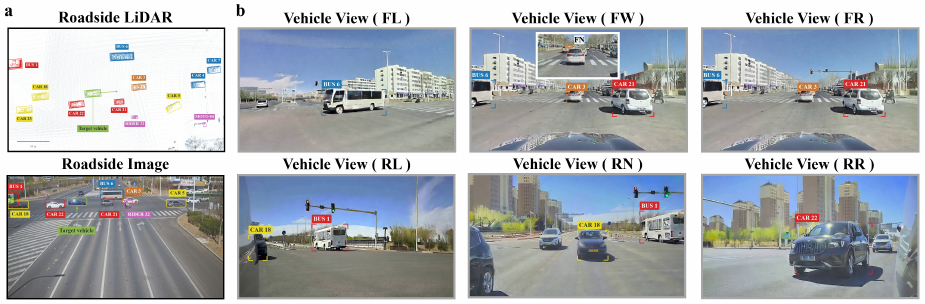}
\caption{\textbf{Reuse of roadside 3D annotations in generated vehicle views.}
\textbf{a} Roadside LiDAR point cloud, camera image and object annotations. Green markers identify the target vehicle.
\textbf{b} Seven camera views generated for the target vehicle and their correspondence with the roadside annotations. Matching colours, categories and identifiers denote the same road users in the roadside observations and generated images. Existing roadside 3D bounding boxes are transformed into the target vehicle's coordinate frame and paired with the generated images. Camera abbreviations follow Fig.~\ref{fig:sequence}; the FN view is inset within the FW view.}
\label{fig:caseA}
\end{figure}

Figure~\ref{fig:caseA} illustrates the mapping of roadside 3D annotations into generated vehicle views. After coordinate transformation, the annotations correspond to the road users in the generated images. The distribution of surrounding vehicles across camera views also agrees with their relative positions in the roadside observations, indicating that the generated images retain the object correspondences and spatial layout of the original traffic scene.

\subsection{Geometric reconstruction quality of dynamic objects}
\label{sec:anyr2v-results-geometry}

To evaluate the geometric conditions used for video generation, we project scenes reconstructed from roadside observations into onboard camera views and compare the rendered depth with motion-compensated onboard LiDAR depth at the corresponding times. We assess both depth accuracy and valid-depth coverage for dynamic objects. The baselines include single-frame roadside point clouds, static-background fusion and methods that predict or supplement geometry using roadside images.

Our method achieves the highest depth accuracy and valid-depth coverage for dynamic objects among the compared methods (Table~\ref{tab:dynamic_geometry}). Its relative depth error (AbsRel) is 0.161, a 76.7\% reduction from the 0.692 obtained by Static-map fusion. Valid-depth coverage reaches 89.24\%, exceeding LiDAR + WorldMirror's 66.76\% by 22.5 percentage points. Our method also achieves the highest depth threshold accuracy ($\delta_1$) and projected-mask intersection over union (Mask IoU). These results show that the reconstructed geometry reduces depth error while extending valid-depth coverage over dynamic objects in onboard views.

\begin{table}[!htbp]
\centering
\caption{\textbf{Quantitative evaluation of roadside-derived geometry in onboard views.} All methods reconstruct scenes from roadside observations and render depth at the same onboard poses and times, using motion-compensated onboard LiDAR depth as the reference. AbsRel and $\delta_1$ are computed at dynamic-object reference pixels where the method being evaluated provides valid depth. Valid pixels denotes valid-depth coverage. Mask IoU measures the intersection over union between projected object masks and segmentation masks predicted from real onboard images within a common reference region. Arrows indicate the preferred direction for each metric; bold and underlined values denote the best and second-best results, respectively.}

\label{tab:dynamic_geometry}
\begingroup
\TableRefineSetup{9}
\begin{tabularx}{\linewidth}{l *{4}{C}}
\toprule
\TableHead{Method} & \mbox{AbsRel\,$\downarrow$} & \mbox{$\delta_1$\,\TableUnit{(\%)}\,$\uparrow$} & \mbox{Mask IoU\,\TableUnit{(\%)}\,$\uparrow$} & \mbox{Valid pixels\,\TableUnit{(\%)}\,$\uparrow$} \\
\midrule
\rowcolor{TableHeaderSub}
\multicolumn{5}{l}{\cellcolor{TableHeaderSub}\TableGroupTag{LiDAR baselines}} \\
\hspace*{8pt}Single-frame LiDAR & 1.050 & 51.27 & 30.84 & 61.43 \\
\hspace*{8pt}Aligned LiDAR & 0.961 & 57.98 & 36.03 & 62.48 \\
\hspace*{8pt}Static-map fusion & \TableRunnerUp{0.692} & \TableRunnerUp{62.11} & 33.56 & 54.65 \\
\addlinespace[3pt]
\rowcolor{TableHeaderSub}
\multicolumn{5}{l}{\cellcolor{TableHeaderSub}\TableGroupTag{Image-based models}} \\
\hspace*{8pt}Depth Anything 3~\cite{lin2026da3} & 1.089 & 35.37 & 21.30 & 58.52 \\
\hspace*{8pt}WorldMirror~\cite{liu2026worldmirror} & 0.704 & 47.71 & 24.18 & 51.90 \\
\addlinespace[3pt]
\rowcolor{TableHeaderSub}
\multicolumn{5}{l}{\cellcolor{TableHeaderSub}\TableGroupTag{Fusion-based representations}} \\
\hspace*{8pt}LiDAR + WorldMirror~\cite{liu2026worldmirror} & 0.905 & 56.21 & \TableRunnerUp{37.42} & \TableRunnerUp{66.76} \\
\midrule
\rowcolor{TableSoftBlue}
\hspace*{8pt}\textbf{OURS} & \TableBestCell{0.161} & \TableBestCell{90.51} & \TableBestCell{45.31} & \TableBestCell{89.24} \\
\bottomrule
\end{tabularx}
\endgroup
\end{table}

\begin{figure}[!htbp]
\centering
\includegraphics[width=\textwidth]{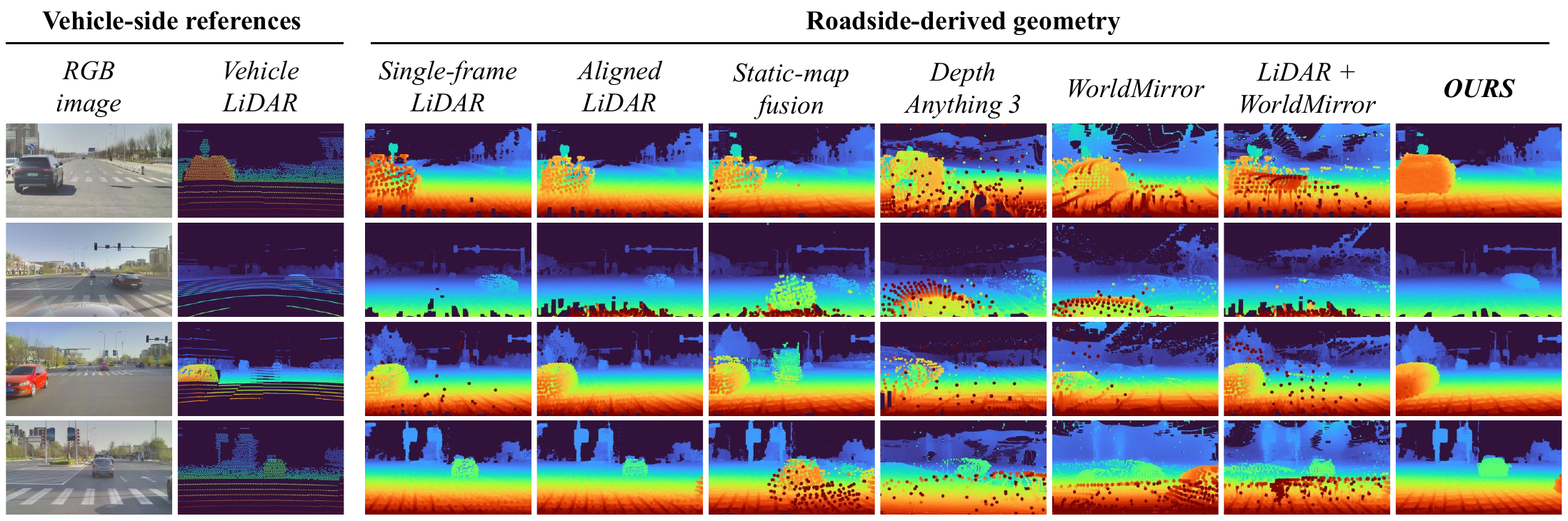}
\caption{\textbf{Depth comparison of roadside scene representations from onboard viewpoints.} Each row shows an example onboard view. The first two columns show the real onboard image and the onboard LiDAR depth reference; the remaining columns show depth rendered from scenes reconstructed by the different methods. Colours encode inverse depth, with red indicating nearby surfaces and blue indicating distant surfaces. A common colour scale is used within each row.}
\label{fig:depth}
\end{figure}

Figure~\ref{fig:depth} compares depth rendered from different roadside scene representations at onboard viewpoints. Single-frame point clouds and static-background fusion exhibit pronounced sampling gaps in vehicle regions, with background depth visible inside vehicle silhouettes. Our method produces more continuous vehicle surfaces and sharper depth boundaries between objects and the background, consistent with the improved valid-depth coverage in Table~\ref{tab:dynamic_geometry}.

\subsection{Effects of tri-modal conditioning on video generation quality}
\label{sec:anyr2v-results-controls}

We assess the contributions of appearance, depth and layout conditions by disabling each modality in turn during inference with the same model. All other inputs are held fixed, and the single-modality ablations are compared with configurations that retain all three conditions or disable them simultaneously. The full tri-modal configuration achieves the best PSNR, SSIM and LPIPS, whereas disabling all three conditions yields the lowest performance (Table~\ref{tab:control}).

\begin{table}[!htbp]
\centering
\caption{\textbf{Effects of appearance, depth and layout conditions on video generation quality.} All configurations use the same model and paired test videos. A check mark indicates that the corresponding condition is retained; a dash indicates that its encoded latent representation is set to zero. Arrows indicate the preferred direction for each metric; bold and underlined values denote the best and second-best results, respectively.}
\label{tab:control}
\begingroup
\TableSetup{9}
\setlength{\aboverulesep}{0pt}\setlength{\belowrulesep}{0pt}
\renewcommand{\TableBestCell}[1]{\textbf{#1}}
\renewcommand{\TableRunnerUp}[1]{\underline{#1}}
\begin{tabularx}{\linewidth}{*{3}{>{\hsize=.85\hsize\linewidth=\hsize}C} *{3}{>{\hsize=1.15\hsize\linewidth=\hsize}C}}
\toprule
\multicolumn{3}{c}{\TableGroup{Control inputs}} & \multicolumn{3}{c}{\TableGroup{Generation quality}} \\
\cmidrule(lr){1-3}\cmidrule(lr){4-6}
\rowcolor{TableHeaderSub}
Appearance & Depth & Layout & \mbox{PSNR\,\TableUnit{(dB)}\,$\uparrow$} & \mbox{SSIM\,$\uparrow$} & \mbox{LPIPS\,$\downarrow$} \\
\midrule
$\checkmark$ & $\checkmark$ & $\checkmark$ & \TableBestCell{15.240} & \TableBestCell{0.4098} & \TableBestCell{0.3675} \\
$\checkmark$ & $\checkmark$ & --- & 14.886 & 0.3985 & 0.4261 \\
--- & $\checkmark$ & $\checkmark$ & 13.144 & 0.2946 & 0.6242 \\
$\checkmark$ & --- & $\checkmark$ & \TableRunnerUp{14.995} & \TableRunnerUp{0.4056} & \TableRunnerUp{0.3894} \\
--- & --- & --- & 12.466 & 0.2808 & 0.6931 \\
\bottomrule
\end{tabularx}
\endgroup
\end{table}

Among the single-modality ablations, disabling appearance produces the largest loss of fidelity, increasing LPIPS by 69.9\% and reducing PSNR by 2.10\,dB relative to the full tri-modal configuration. With depth and layout retained, the generated images still differ substantially from the real images, highlighting the importance of roadside colour and texture observations for matching the generated videos to the recorded scene.

Disabling layout or depth alone increases LPIPS by 15.9\% and 6.0\%, respectively. Appearance projections already encode the image-space positions of scene surfaces and objects, so the remaining conditions retain some spatial information when either depth or layout is disabled. Explicit depth and layout conditions further improve fidelity, demonstrating the complementary roles of the three modalities in representing scene appearance and spatial structure.

\subsection{Training benefits of generated data for 3D object detection}
\label{sec:anyr2v-results-detection}

To evaluate the training value of the generated data for 3D object detection, we generate surround-view videos along other vehicles' trajectories using roadside recordings from the training set. Roadside 3D annotations are transformed into the corresponding vehicle coordinate frames to provide labelled training samples. We then train BEVFormer, StreamPETR and CAPE with different amounts of generated data added to the real training set and evaluate detection accuracy across six object categories on the same real onboard test set.

Adding generated frames equivalent to 50\% of the original real training frame count improves mAP on the real onboard test set by 3.3, 3.3 and 3.4 AP-points for BEVFormer, StreamPETR and CAPE, respectively, relative to their real-data-only baselines (Table~\ref{tab:detection_core}a). These gains show that data generated from roadside recordings along other vehicles' trajectories provide effective training samples for 3D object detection.

\begin{table}[!htbp]
\centering
\caption{\textbf{Effects of generated data on real-world onboard 3D object detection.} Different amounts of generated data are added to the real training set, and all configurations are evaluated on the same real onboard test set. Added denotes the ratio of added generated frames to the original number of real training frames, expressed as a percentage; 0\% denotes real-data-only training.
\textbf{a} Six-class mAP at different generated-data addition ratios.
\textbf{b} AP for motorcycles, bicycles and barriers at a 50\% addition ratio. Base and Aug. denote real-data-only training and training augmented with generated data, respectively. AP and mAP are expressed as percentages. Values next to the red arrows indicate relative gains over the corresponding baselines. Bold and light blue highlight the highest mAP for each detector across the tested addition ratios.}
\label{tab:detection_core}
\begingroup
\TableRefineSetup{9}
\setlength{\tabcolsep}{1.5pt}
\renewcommand{\arraystretch}{1.2}
\vbox{\offinterlineskip
\hbox{\begin{tabularx}{\linewidth}{>{\centering\arraybackslash}p{27pt} *{3}{C} >{\centering\arraybackslash}p{27pt} *{3}{C}}
\toprule
\multicolumn{8}{l}{\rule[-3pt]{0pt}{18pt}\TablePanel{a}{Six-class mAP (\%)}} \\
\rowcolor{TableHeaderSub}
\TableHead{Added} & \TableHead{BEVFormer} & \TableHead{StreamPETR} & \TableHead{CAPE} & \TableHead{Added} & \TableHead{BEVFormer} & \TableHead{StreamPETR} & \TableHead{CAPE} \\
\midrule
\TableDetectionStrut 0\% & \TableScoreGain{44.5}{} & \TableScoreGain{42.0}{} & \TableScoreGain{44.3}{} & \cellcolor{TableSoftBlue}\textbf{50\%} & \cellcolor{TableSoftBlue}\renewcommand{\TableGainInk}{TableGainStrong}\TableScoreGain{\textbf{47.8}}{7.42\%} & \cellcolor{TableSoftBlue}\renewcommand{\TableGainInk}{TableGainStrong}\TableScoreGain{\textbf{45.3}}{7.86\%} & \cellcolor{TableSoftBlue}\renewcommand{\TableGainInk}{TableGainStrong}\TableScoreGain{\textbf{47.7}}{7.67\%} \\
\TableDetectionStrut 10\% & \TableScoreGain{44.6}{0.22\%} & \TableScoreGain{42.6}{1.43\%} & \TableScoreGain{46.4}{4.74\%} & 60\% & \TableScoreGain{45.9}{3.15\%} & \TableScoreGain{45.0}{7.14\%} & \TableScoreGain{45.5}{2.71\%} \\
\TableDetectionStrut 20\% & \TableScoreGain{46.0}{3.37\%} & \TableScoreGain{43.4}{3.33\%} & \TableScoreGain{46.9}{5.87\%} & 100\% & \TableScoreGain{45.8}{2.92\%} & \TableScoreGain{45.1}{7.38\%} & \TableScoreGain{46.2}{4.29\%} \\
\midrule
\end{tabularx}}
\hbox{\begin{tabularx}{\linewidth}{>{\raggedright\arraybackslash}p{57pt} *{6}{C}}
\multicolumn{7}{l}{\rule[-3pt]{0pt}{18pt}\TablePanel{b}{Category AP (\%) at 50\% addition}} \\
\cmidrule(lr){2-3}\cmidrule(lr){4-5}\cmidrule(lr){6-7}
\addlinespace[2pt]
 & \multicolumn{2}{c}{\TableHead{BEVFormer}} & \multicolumn{2}{c}{\TableHead{StreamPETR}} & \multicolumn{2}{c}{\TableHead{CAPE}} \\
\cmidrule(lr){2-3}\cmidrule(lr){4-5}\cmidrule(lr){6-7}
\rowcolor{TableHeaderSub}
\TableHead{Category} & Base & Aug. & Base & Aug. & Base & Aug. \\
\midrule
\TableDetectionStrut Motorcycle & 37.9 & \TableScoreGain{50.2}{32.45\%} & 35.0 & \TableScoreGain{37.7}{7.71\%} & 37.4 & \TableScoreGain{47.4}{26.74\%} \\
\TableDetectionStrut Bicycle & 47.9 & \TableScoreGain{54.9}{14.61\%} & 46.1 & \TableScoreGain{47.2}{2.39\%} & 49.8 & \TableScoreGain{52.0}{4.42\%} \\
\TableDetectionStrut Barrier & 70.0 & \TableScoreGain{76.0}{8.57\%} & 67.2 & \TableScoreGain{68.3}{1.64\%} & 72.7 & \TableScoreGain{74.8}{2.89\%} \\
\bottomrule
\end{tabularx}}
}
\endgroup
\end{table}

Detection gains vary with the proportion of generated data added, with distinct trends across models. Among the tested addition ratios, all three detectors achieve their highest mAP at 50\%. At 100\%, StreamPETR remains close to its peak, whereas BEVFormer and CAPE show more pronounced declines. All configurations combining real and generated training data outperform their respective real-data-only baselines (Table~\ref{tab:detection_core}a). These results show that adding generated data improves detection performance, with the magnitude of the benefit depending on both the addition ratio and detector architecture.

\FloatBarrier

\section{Discussion}
\label{sec:anyr2v-discussion}

AnyR2V transforms multimodal roadside observations into continuous onboard surround-view videos and reuses roadside 3D annotations to construct training data for autonomous driving. R2V360 provides spatiotemporally aligned roadside and real onboard data, enabling paired supervision for model training and quantitative comparison between generated and recorded images. Augmenting real training data with generated samples improves the accuracy of 3D object detectors on the real onboard test set, further demonstrating the practical training value of these samples. Together, these results show that a single roadside recording can yield surround-view videos with corresponding 3D annotations from multiple vehicle perspectives, allowing the same real traffic sequence to be reused across viewpoints. Traffic records continuously accumulated by road infrastructure thus provide a complementary source of autonomous driving training data, with the potential to reduce reliance on repeated vehicle-based collection and annotation.

The present evaluation focuses on an urban intersection. Extending data collection and evaluation to other road environments would help establish how variations in road structure, traffic organisation and roadside sensor layouts affect generation quality and training gains. For dynamic objects, the current instance geometry, fixed within each sequence, could be extended to pose-dependent representations that incorporate limb motion and local deformation of pedestrians and riders into scene modelling and generation conditions. This would enable more accurate depiction of non-rigid motion in continuous surround-view videos. Further evaluation could examine the contribution of generated data to end-to-end driving model training and closed-loop tasks. These directions would help define the applicability of roadside data across traffic environments and training tasks, while improving data generation in complex dynamic scenes.

\section{Methods}
\label{sec:anyr2v-nc2-method}

\subsection{Observations and generation task}
\label{sec:anyr2v-nc2-task}

AnyR2V constructs a shared scene from roadside traffic recordings and generates seven-camera surround-view videos along the recorded vehicle trajectories. The inputs comprise fused point clouds from four LiDAR sensors, corresponding images from four pinhole cameras, and 3D bounding boxes with category labels and tracking identifiers. Point clouds and bounding boxes are expressed in the roadside coordinate frame, and camera intrinsics and extrinsics are known. Multi-frame observations are aggregated and rendered from the target vehicle's camera viewpoints as appearance, depth and layout conditions to guide video generation.

\begin{figure}[!htbp]
\centering
\includegraphics[width=\linewidth]{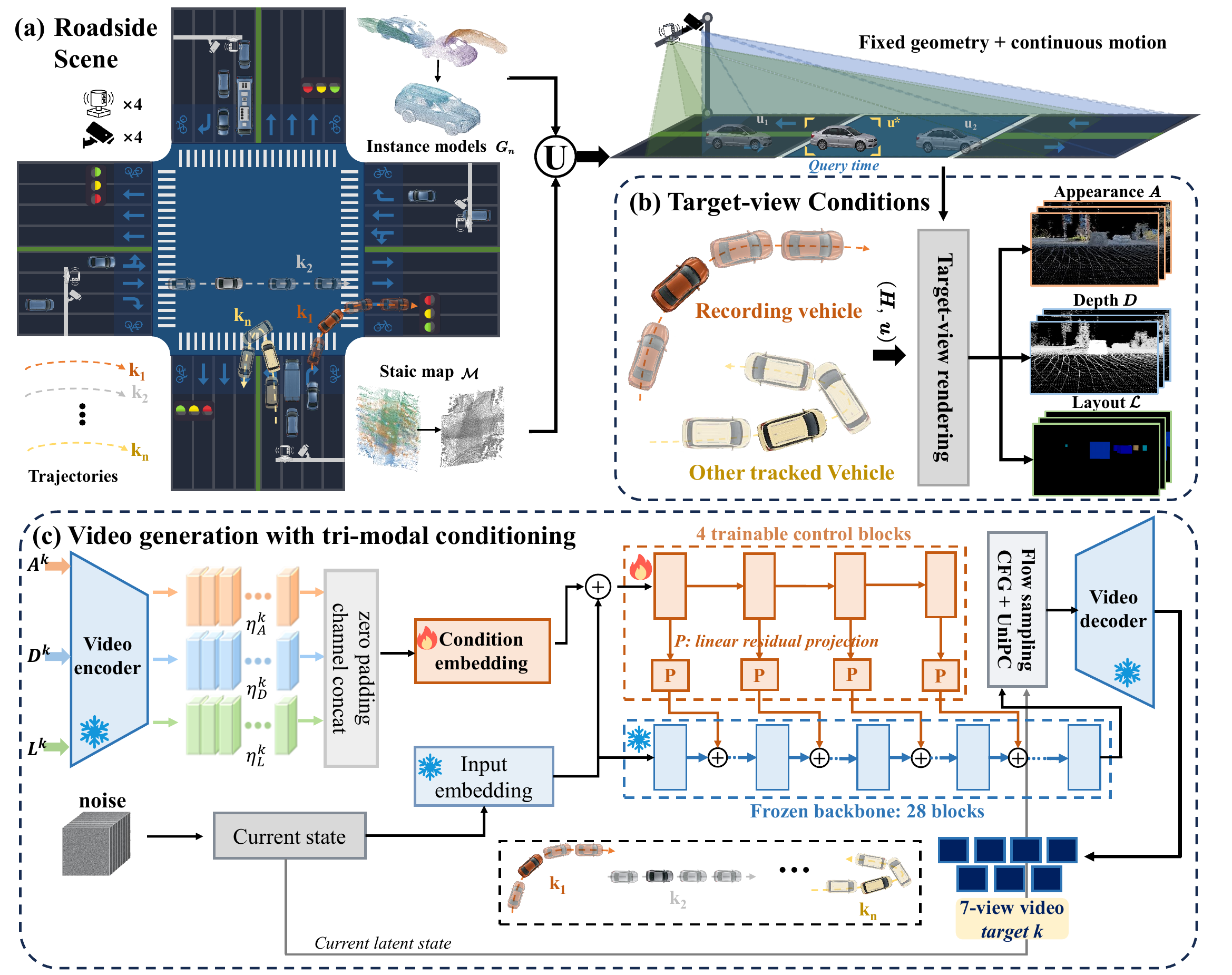}
\caption{\textbf{AnyR2V framework for roadside-to-vehicle surround-view video generation.}
\textbf{a, Construction of the shared roadside scene.} Sequential roadside point clouds and images are aggregated using 3D tracking annotations to construct a static background and instance geometry for dynamic objects. Together with object trajectories, these form a shared scene that can be queried at specified times. Instance geometry remains fixed, while position and orientation vary along each trajectory.
\textbf{b, Rendering of target-view conditions.} Camera poses and capture times are determined along the trajectory of the recording vehicle or another target vehicle. The shared scene is rendered into appearance, depth and layout conditions, which respectively provide observed colours, relative depth distributions, and object categories and spatial extents.
\textbf{c, Video generation with tri-modal conditioning.} The three conditions are encoded by a frozen video encoder and concatenated. A trainable condition embedding and four control blocks inject conditioning residuals into the frozen video backbone, guiding the progressive generation of video latents from Gaussian noise. The seven videos for a given vehicle are generated jointly and decoded separately for each camera by a frozen decoder. Videos for different vehicles are generated independently using the shared scene and model parameters. Flames and snowflakes indicate trainable and frozen modules, respectively. P denotes a linear residual projection; CFG denotes classifier-free guidance, applied here to text conditions; and UniPC is the sampling solver. Text and flow-time inputs and training supervision paths are omitted.}
\label{fig:anyr2v-pipeline}
\end{figure}

Scene geometry, colour and image-based generation conditions all derive from roadside observations. Real surround-view videos from the recording vehicle $e$ supervise model training, and its onboard LiDAR supports localisation and roadside--vehicle spatiotemporal calibration. For other vehicles, virtual camera poses are determined from roadside trajectories, using the same conditioning pipeline and generative model. Each scene is aggregated offline within a single traffic sequence, and videos are generated along trajectories and within valid time intervals covered by the recording. Figure~\ref{fig:anyr2v-pipeline} provides an overview of AnyR2V.

Within each sequence, $k$ denotes the target vehicle, $v\in\mathcal C$ denotes one of the seven camera views, and $i$ indexes video frames. Vehicle and roadside times are denoted by $t$ and $u$, respectively. The roadside coordinate frame is denoted by $r$, and the rigid transform $\mathbf T_{a\leftarrow b}$ maps points from coordinate frame $b$ to coordinate frame $a$.

\subsection{Temporal aggregation of roadside observations}
\label{sec:anyr2v-nc2-scene}

Surface sampling in a single roadside point cloud is sparse, allowing the background to appear through gaps in foreground surfaces when projected into onboard views. To improve surface coverage, we accumulate the static background in the roadside coordinate frame and aggregate pose-aligned observations of moving objects into instance point clouds. Instance geometry and motion trajectories are stored separately, and the scene is reassembled at the queried object poses (Fig.~\ref{fig:anyr2v-pipeline}a).

For instance $n$, we extract the point set $P_{n,m}$ at roadside time $u_m$ from a moderately expanded 3D bounding box. We define a body-fixed coordinate frame with its origin at the box centre, its $x$ axis aligned with the annotated heading and its $z$ axis vertical. The transform from this body frame to the roadside frame is denoted by $\mathbf T_n(u_m)$. Applying its inverse to each frame's point set brings the observations into a common body frame for aggregation. Poses for non-keyframes are interpolated from annotations, and interpolation errors can misalign surfaces across frames. We therefore alternate pose refinement and updates to the aggregated point cloud.

Pose refinement is initialised from the annotations. In each round, observations are assigned cyclically to five groups according to frame index. For each group, a reference point cloud is constructed from the other four groups, and each frame in the selected group is registered to this reference. The observations being registered and the reference thus come from different frame groups. Registration uses the point-to-point iterative closest point (ICP) algorithm~\cite{besl1992registration} to optimise 3D translation and yaw, with progressively tighter correspondence-distance thresholds. If $\mathbf U$ aligns an observation in its current body frame to the reference, the pose is updated as $\mathbf T_n\leftarrow\mathbf T_n\mathbf U^{-1}$. The point sets originally extracted from each frame remain fixed.

The current pose is retained for frames with too few valid correspondences or corrections exceeding the allowed threshold. After two refinement rounds, remaining gaps in the pose corrections are filled by interpolation from neighbouring frames. The observations are then fused using $0.05\,\mathrm{m}$ voxels to obtain the instance point cloud $G_n$. Each instance has fixed geometry throughout the sequence, with bounding-box dimensions taken from its first-frame annotation. This rigid-body approximation also applies to pedestrians and riders.

Points outside the object bounding boxes in each frame are accumulated using $0.1\,\mathrm{m}$ voxels to form the initial static map $\mathcal M_{\mathrm{raw}}$. We then remove residual points using the refined object trajectories, retain voxels observed in at least three distinct frames, and filter out isolated points to obtain the background map $\mathcal M$ used for rendering.

Geometric aggregation also preserves colours observed by the roadside cameras. Original points from each frame are projected into the corresponding images, using a depth buffer constructed from the point cloud of that frame. Each point's projected footprint is adjusted according to its optical-axis depth. For points within the image bounds that pass visibility checks, colours are sampled by bilinear interpolation, and valid colours from multiple cameras are averaged with equal weights.

Colour is fused alongside geometry. For each static voxel, colours are first averaged within each frame, followed by an equally weighted average across frames. Instance voxels pool valid point colours from all frames. Only valid observations contribute to colour averages; geometry is retained for points with missing colour. The fused colour, denoted by $a(\mathbf p)$, is stored with each point for subsequent scene queries.

At query time, instance poses are evaluated at the camera capture time. Between adjacent roadside sampling times, positions are linearly interpolated and yaw angles are interpolated after unwrapping. Let $\mathcal N(u)$ denote the set of instances with valid states at roadside time $u$. The shared scene is
\begin{equation}
\mathcal S(u)
=\mathcal M\cup\bigcup_{n\in\mathcal N(u)}\mathbf T_n(u)G_n.
\label{eq:anyr2v-nc2-scene}
\end{equation}
Multi-frame aggregation supplies the background and instance geometry within each sequence, while changes in instance poses capture the temporal evolution of the scene, enabling queries throughout the observation interval.

\subsection{Spatiotemporal calibration between roadside sensors and the recording vehicle}
\label{sec:anyr2v-nc2-alignment}

To align the shared scene with training videos from the recording vehicle, we estimate the vehicle's trajectory in the roadside coordinate frame and calibrate the temporal relationship between the two platforms (Fig.~\ref{fig:overview}c). The initial trajectory, denoted by $\widehat{\mathbf T}_e(t)$, is obtained from onboard LiDAR odometry and aligned to the roadside frame through static-scene registration. This transform maps onboard LiDAR coordinates at time $t$ into the roadside frame. Between sampled times, we use linear interpolation for translation and spherical interpolation for rotation.

Spatial correction aligns the cameras with the scene, and temporal correction aligns moving objects with the camera capture times. Within each sequence, the roadside--vehicle time offset is assumed to be a constant $\Delta$, such that vehicle time $t$ corresponds to roadside time $t+\Delta$. A six-degree-of-freedom rigid transform $\mathbf T_{\mathrm{corr}}(\boldsymbol{\xi})$, shared across the sequence, left-multiplies the initial trajectory $\widehat{\mathbf T}_e(t)$ to correct its global misalignment in the roadside frame. Here, $\boldsymbol{\xi}$ contains the rotation and translation parameters. Joint estimation optimises $\Delta$ and $\boldsymbol{\xi}$ while holding all instance trajectories in the shared scene and the initial recording-vehicle trajectory fixed.

Calibration uses onboard LiDAR observations with per-point timestamps. Each scan is divided into short time intervals, and the interpolated pose at each interval centre is used to compensate for recording-vehicle motion. This yields points $\mathbf X_j$ in the roadside frame while preserving their actual acquisition times $t_j$. Dynamic points are assigned to instances using expanded object extents under a preset time offset, and these assignments remain fixed during optimisation. Instance poses are queried at $t_j+\Delta$, allowing the time offset to affect the residuals through object motion.

For a dynamic point assigned to instance $n_j$, we apply the spatial correction and transform the point into the instance's body frame at $t_j+\Delta$ before comparing it with the corresponding point $\mathbf q_j^d$ in the instance point cloud. Static points are compared with surfaces near the recording-vehicle trajectory in the initial map $\mathcal M_{\mathrm{raw}}$, with corresponding points and local normals denoted by $\mathbf q_j^s$ and $\mathbf n_j$, respectively. We use point-to-point residuals for dynamic points and point-to-plane residuals for static points:
\begin{equation}
\begin{aligned}
\mathbf r_j^d(\Delta,\boldsymbol{\xi})
&=\mathbf T_{n_j}(t_j+\Delta)^{-1}\mathbf T_{\mathrm{corr}}(\boldsymbol{\xi})\mathbf X_j-\mathbf q_j^d,\\
r_j^s(\boldsymbol{\xi})
&=\mathbf n_j^\top\left(\mathbf T_{\mathrm{corr}}(\boldsymbol{\xi})\mathbf X_j-\mathbf q_j^s\right).
\end{aligned}
\label{eq:anyr2v-nc2-residuals}
\end{equation}
These residuals are evaluated in the instance body frames and the roadside frame, respectively. Nearest-neighbour correspondences are recomputed at each outer iteration and held fixed during that iteration's parameter update. Correspondence searches for dynamic points are restricted to the assigned instance.

To reduce the influence of outlier correspondences, we apply the Geman--McClure robust function $\rho_\beta(a)=a/(a+\beta^2)$ to squared residuals, where $\beta$ controls the residual scale. Let $\mathcal I_d$ and $\mathcal I_s$ denote the index sets of valid dynamic and static points, respectively. Averaging each cost term separately gives the joint objective
\begin{equation}
\begin{aligned}
E(\Delta,\boldsymbol{\xi})
={}&\frac{1}{|\mathcal I_d|}\sum_{j\in\mathcal I_d}
\rho_{\beta_d}\!\left(\|\mathbf r_j^d(\Delta,\boldsymbol{\xi})\|_2^2\right)\\
&+\frac{1}{|\mathcal I_s|}\sum_{j\in\mathcal I_s}
\rho_{\beta_s}\!\left([r_j^s(\boldsymbol{\xi})]^2\right).
\end{aligned}
\label{eq:anyr2v-nc2-alignment-objective}
\end{equation}
The dynamic and static residual scales are set to $\beta_d=0.15\,\mathrm{m}$ and $\beta_s=0.10\,\mathrm{m}$, respectively. We initialise $\Delta$ using a coarse-to-fine grid search, then alternate correspondence updates with a fixed number of gradient-based optimisation iterations to obtain $\widehat\Delta$ and $\widehat{\boldsymbol{\xi}}$. Static surfaces constrain the global spatial correction, while changes in dynamic-object positions over time constrain the temporal offset. Together, they establish spatiotemporal correspondence between the training videos and the shared scene.

\subsection{Construction of target-view conditioning videos}
\label{sec:anyr2v-nc2-rendering}

Following spatiotemporal calibration, conditions are rendered using the target camera pose and corresponding roadside time (Fig.~\ref{fig:anyr2v-pipeline}b). Let $\mathbf H_i^{k,v}$ denote the transform from camera $v$ of vehicle $k$ to the roadside coordinate frame at frame $i$, and let $u_i^{k,v}$ denote the corresponding roadside time. For the recording vehicle, the fixed extrinsic transform $\mathbf E_v$ maps camera coordinates to onboard LiDAR coordinates. Assuming that the onboard cameras and LiDAR share a time base, the trajectory is queried at each camera's own capture time $t_i^{e,v}$.

For another vehicle $k$, the virtual camera rig follows the refined roadside trajectory $\mathbf T_k(u)$. Its mounting reference frame has its origin $0.25\,\mathrm{m}$ above the centre of the bounding box's top face, with axes aligned with the vehicle body. The transform from this mounting frame to the vehicle's body frame is denoted by $\mathbf B_k$. The cameras reuse the extrinsic transforms $\mathbf E_v$ and the intrinsics of the recording vehicle. Let $u_i^k$ denote an output time on the roadside clock for another vehicle. The camera poses and scene query times are then
\begin{equation}
\left(\mathbf H_i^{k,v},u_i^{k,v}\right)
=\begin{cases}
\left(\mathbf T_{\mathrm{corr}}(\widehat{\boldsymbol{\xi}})\widehat{\mathbf T}_e(t_i^{e,v})\mathbf E_v,\;t_i^{e,v}+\widehat\Delta\right),&k=e,\\[3pt]
\left(\mathbf T_k(u_i^k)\mathbf B_k\mathbf E_v,\;u_i^k\right),&k\ne e.
\end{cases}
\label{eq:anyr2v-nc2-camera-query}
\end{equation}
The recording-vehicle trajectory is queried on the vehicle clock, and $\widehat\Delta$ determines the corresponding roadside scene time. Other vehicles' trajectories and the scene share the roadside time axis. The camera rig's mounting height adapts to vehicle height, while its internal configuration remains fixed.

Before rendering, the target vehicle's own instance point cloud and layout box are removed from the queried scene. Let $K_v$ denote the camera intrinsics after undistortion and resizing to the output resolution. A scene point $\mathbf p$ projects to the pixel location
\begin{equation}
\mathbf x_i^{k,v}(\mathbf p)
=\pi_{K_v}\!\left((\mathbf H_i^{k,v})^{-1}\mathbf p\right),
\label{eq:anyr2v-nc2-projection}
\end{equation}
where $\pi_{K_v}$ denotes pinhole projection, and the third component of the camera coordinates gives the optical-axis depth. Supervision images and conditioning images use the same undistorted image plane and output dimensions to preserve pixelwise correspondence.

Appearance and depth are produced in a single point-cloud rendering pass. Each point covers a small neighbourhood around its projected location, and the point with the smallest optical-axis depth determines the colour and depth at overlapping pixels. For a given image, let $\mathcal P(\mathbf x)$ denote the set of points covering pixel $\mathbf x$, and let $z(\mathbf p)$ denote a point's optical-axis depth. For non-empty $\mathcal P(\mathbf x)$,
\begin{equation}
\begin{aligned}
\mathbf p^\star(\mathbf x)
&=\arg\min_{\mathbf p\in\mathcal P(\mathbf x)}z(\mathbf p),\\
Z(\mathbf x)&=z\!\left(\mathbf p^\star(\mathbf x)\right),\\
A(\mathbf x)&=\begin{cases}
a\!\left(\mathbf p^\star(\mathbf x)\right),&\text{valid colour},\\
0,&\text{missing colour}.
\end{cases}
\end{aligned}
\label{eq:anyr2v-nc2-rasterization}
\end{equation}
Foreground points with missing colour still determine depth at the pixels they cover, with appearance set to zero to preserve foreground occlusion. Uncovered pixels are assigned zero appearance and depth.

The depth condition $D$ supplied to the encoder is derived from metric depth $Z$. For each frame and camera, the valid depth range is linearly mapped to greyscale, with nearby surfaces brighter and distant surfaces darker. The greyscale map is then replicated across three channels. This per-image normalisation makes intensity encode the relative depth distribution within each image. Occlusion is evaluated using metric depth.

The layout condition $L$ is rendered from instance poses and bounding-box dimensions at the same time. The 3D boxes are represented as solid cuboids coloured by category, with face shading providing orientation cues. Layout rendering is independent of the scene point-cloud depth buffer; boxes are composited from far to near according to their mean optical-axis depth. At the object level, layout preserves categories and spatial extents, supplying structural information in sparsely sampled regions. Appearance and depth conditions retain the surface observations from the point cloud.

Rendering frame by frame along the target trajectory yields conditioning videos $C^k=(A^k,D^k,L^k)$. All three modalities share camera parameters, scene times and target trajectories, aligning surface observations and object layouts at corresponding image locations for joint input to the video generation network.

\subsection{Surround-view video generation with tri-modal conditioning}
\label{sec:anyr2v-nc2-generation}

These conditions organise roadside observations in the target views, and the video generation model completes image content in regions with sparse surface samples or missing texture. Building on a pretrained multi-view video model~\cite{WorldSimulationWithVideoFoundationModels}, we fine-tune the control branch that processes these conditions to learn a mapping from target-view conditions to real onboard videos, jointly generating the seven views for each vehicle (Fig.~\ref{fig:anyr2v-pipeline}c).

Each training sample contains 29 real video frames per camera for seven cameras and the corresponding tri-modal conditions, all at a resolution of $1280\times720$. A frozen video encoder $\mathcal E$ independently encodes each camera's video and conditions before stacking them in a fixed view order. For a given vehicle, let $V^v$ denote the real video from camera $v$, $z_0$ the joint video latent and $\eta_X$ the conditioning latent for modality $X$. Then
\begin{equation}
\begin{aligned}
z_0&=\operatorname{Stack}_{v\in\mathcal C}\mathcal E(V^v),\\
\eta_X&=\operatorname{Stack}_{v\in\mathcal C}\mathcal E(X^v),\qquad X\in\{L,A,D\},\\
\eta&=\operatorname{Pad}_0\!\left([\eta_L;\eta_A;\eta_D]_{\mathrm{ch}}\right).
\end{aligned}
\label{eq:anyr2v-nc2-latents}
\end{equation}
Here, $\operatorname{Stack}$ arranges the views along the latent temporal dimension, and $[\cdot]_{\mathrm{ch}}$ denotes channel-wise concatenation. Layout, appearance and depth each occupy 16 channels, forming a 48-channel condition in that order, which is zero-padded to the control interface's 184 channels. The layout channels occupy the positions reserved for structured-rendering inputs in the pretrained configuration, followed by the appearance and depth channels. The control branch is initialised with pretrained weights.

Videos and conditions share the same spatiotemporal grid and view order. Each camera sequence is encoded independently, with temporal positions indexed from the start of that sequence and camera embeddings identifying the view. This positional information is retained after stacking to support cross-view interaction within the generation network.

Each camera also receives a fixed textual description of its viewpoint. For the front wide-angle camera, the text additionally includes a templated description of the environment and the driving direction. The direction is taken from the recorded annotations for the recording vehicle and determined from roadside trajectories for other vehicles. A frozen text encoder produces features $d$, which condition generation through within-view cross-attention.

Video generation uses a rectified-flow parameterisation~\cite{liu2023rectifiedflow}, in which flow time $\tau\in[0,1]$ controls interpolation between the video latent and Gaussian noise:
\begin{equation}
z_\tau=(1-\tau)z_0+\tau\epsilon,
\qquad\epsilon\sim\mathcal N(0,I).
\label{eq:anyr2v-nc2-flow-path}
\end{equation}
The data and noise endpoints correspond to $\tau=0$ and $\tau=1$, respectively. Flow time parametrises the generation process and is defined separately from the physical time of the traffic recording. All views and frames within a sample share $\tau$, while noise is sampled independently for each latent element. The visual condition $\eta$ retains its encoded value.

To predict velocity along the flow path, the current video latent $z_\tau$ and conditioning latent $\eta$ are separately mapped to feature tensors of identical shape with a feature dimension of 2048. Video features enter the frozen backbone. After input projection, condition features are added elementwise to video features at the input to the first control block. The four control blocks process features sequentially, incorporating flow time and text features. Linear projections of their outputs produce residuals that are added to the outputs of layers 1, 8, 15 and 22 of the 28-layer backbone, respectively. Processing continues through layer 28, followed by an output projection that yields the velocity prediction. Self-attention in both the backbone and control branch spans all views, latent time steps and spatial positions, jointly processing the seven videos. With $\theta$ denoting the trainable control-branch parameters, the full network output is $v_\theta(z_\tau,\tau,\eta,d)$, which has the same shape as $z_\tau$.

Training updates the control branch's condition embedding, prompt-feature mapping and four control blocks, including their input and residual projections. The video input projection, backbone and output projection, video encoder and decoder, text encoder and camera embeddings remain frozen. The precomputed scenes and calibration parameters are also fixed during generative model training. Real videos are encoded as $z_0$, and the derivative $\epsilon-z_0$ of the flow path in Eq.~\eqref{eq:anyr2v-nc2-flow-path} serves as the velocity target. Mean squared error is computed with equal weight across all latent elements. To enable text guidance at inference, all text features are set to zero with probability $0.2$ during training, while all three visual conditions are retained. Let $b\sim\operatorname{Bernoulli}(0.8)$ and let $N_z$ denote the total number of latent elements. The training objective is
\begin{equation}
\mathcal L(\theta)
=\mathbb E_{(z_0,\eta,d),\epsilon,\tau,b}
\left[\frac{1}{N_z}
\left\|v_\theta(z_\tau,\tau,\eta,bd)-(\epsilon-z_0)\right\|_2^2\right].
\label{eq:anyr2v-nc2-training}
\end{equation}
Flow time is sampled from a logit-normal distribution, then discretised and shifted according to the prescribed configuration.

At inference, the seven-camera video latent is initialised from standard Gaussian noise, and sampling is guided by the target vehicle's visual conditions $\eta^k$ and auxiliary text $d^k$. Sampling is independent of real initial frames or prefix videos from the target vehicle. Consistent with training-time text dropout, the network evaluates $v_c(z,\tau)=v_\theta(z,\tau,\eta^k,d^k)$ with text retained and $v_0(z,\tau)=v_\theta(z,\tau,\eta^k,0)$ with text features set to zero, using the same visual conditions in both cases. The guided velocity is
\begin{equation}
v_{\mathrm g}(z,\tau)
=v_c(z,\tau)+\omega\bigl[v_c(z,\tau)-v_0(z,\tau)\bigr],
\label{eq:anyr2v-nc2-guidance}
\end{equation}
where $\omega$ controls the strength of text guidance. At each step, the current latent and guided velocity yield the clean-latent estimate $\widetilde z_0^{(i)}=z_{\tau_i}-\tau_i v_{\mathrm g}(z_{\tau_i},\tau_i)$ for the second-order UniPC predictor--corrector update~\cite{zhao2023unipc}. Sampling proceeds through 35 updates from the noise endpoint towards the data endpoint, keeping the visual conditions $\eta^k$ fixed and using the updated latent for the next velocity prediction. The final state $\widehat z_0$ is split by camera, and the frozen decoder restores the latent scale and decodes the seven videos separately.

All target vehicles share the trained model parameters. The seven videos for a given vehicle are generated jointly, with separate sampling for different vehicles. To switch target vehicles, we reuse the shared scene, determine camera poses and query times along the new target trajectory, remove the target vehicle's own instance and render its conditions. The model then generates videos from that vehicle's viewpoints.

\begingroup
\color{black}
\ifdefined\hypersetup
  \hypersetup{linkcolor=black,citecolor=black,urlcolor=black}
\fi

\subsection{Experimental settings and evaluation}
\label{sec:anyr2v-evaluation}
\label{sec:anyr2v-eval-data}
\label{sec:anyr2v-eval-video}
\label{sec:anyr2v-eval-geometry}
\label{sec:anyr2v-eval-controls}
\label{sec:anyr2v-eval-other-vehicles}
\label{sec:anyr2v-eval-detection}

The 89 recorded sequences in R2V360 are divided into training, validation and test sets containing 71, 9 and 9 sequences, respectively. The training set is used to train the generative model, the validation set for hyperparameter tuning and model selection, and the test set to evaluate video generation and compare tri-modal conditioning configurations.

Across all test sequences, we compare generated videos with synchronously recorded onboard videos from all seven camera views. Both sets of images are aligned on the same undistorted image plane at a resolution of $1280\times720$. The baselines are DropGaussian~\cite{wu2024dropgaussian}, \dtgs{}~\cite{song2026d2gs}, Co-Adaptation~\cite{zhou2025coadaptation}, SparseGS~\cite{li2024sparsegs}, AD-GS~\cite{hu2024adgs} and S3Gaussian~\cite{huang2024s3gaussian}. Each method reconstructs the scene from the four roadside pinhole camera views, using the fused roadside point cloud for initialisation, then renders images at the corresponding onboard camera poses. All methods are evaluated against the same set of onboard reference images.

We use PSNR, SSIM~\cite{wang2004ssim} and LPIPS~\cite{zhang2018lpips} to assess the fidelity of generated images to real onboard images, and KID~\cite{binkowski2018kid} and FID~\cite{heusel2017fid} to measure differences between the two image distributions. In Table~\ref{tab:gs}, KID and FID are computed over image sets, and the remaining metrics are averaged over per-image results. KID values are reported after multiplication by $10^3$.

To assess geometric consistency between generated images and the real scene, we use the same Depth Anything V2 model~\cite{yang2024depthanythingv2} to predict relative depth for generated and real onboard images, exclude distant regions such as the sky, and perform scale-and-shift alignment in disparity space. Let $d$ and $d^\ast$ denote the aligned depth predictions for generated and real images, respectively. The relative depth error AbsRel is the mean of $|d-d^\ast|/d^\ast$, and the depth threshold accuracy $\delta_1$ is the proportion of pixels satisfying $\max(d/d^\ast,d^\ast/d)<1.25$. Semantic consistency is measured by Seg-Cons, the pixelwise agreement between semantic labels predicted for both image sets by the same Mask2Former model~\cite{cheng2022mask2former}.

For target vehicles lacking synchronously recorded onboard videos, roadside images, point clouds and 3D annotations provide references for comparing road layouts, relative vehicle positions and motion relationships in the generated images.

We evaluate roadside-derived geometric conditions using the training sequences. Each method renders depth at the same onboard camera poses and times for comparison with the corresponding onboard LiDAR observations. Motion-compensated onboard LiDAR provides the reference depth.

The geometric baselines cover single-frame point clouds, static-background fusion and image-based geometry prediction. Single-frame LiDAR uses a single roadside point cloud with the original timestamp pairing. Aligned LiDAR adds spatiotemporal alignment and compensates dynamic points to the query time using object trajectories. Static-map fusion combines a multi-frame static background with single-frame dynamic points. Depth Anything 3~\cite{lin2026da3} and WorldMirror~\cite{liu2026worldmirror} use the four roadside pinhole images from the same time and known camera poses. WorldMirror additionally uses camera intrinsics and roadside LiDAR to determine the scale of the predicted geometry. LiDAR + WorldMirror fuses the scale-aligned predicted point cloud with a single-frame roadside point cloud. Our method constructs scene geometry through the temporal aggregation of roadside observations described in Section~\ref{sec:anyr2v-nc2-scene}.

Depth accuracy of the geometric conditions is evaluated using AbsRel and $\delta_1$, with reference depth $d^\ast$ obtained from onboard LiDAR. Both metrics are computed at dynamic-object reference pixels where the method being evaluated provides valid depth. Valid-depth coverage, reported as Valid pixels, is the proportion of all dynamic-object reference pixels with finite, positive predicted depth. Mask IoU compares object masks projected from the scene geometry with those extracted from real onboard images by Mask2Former within a common reference region. Each metric is first averaged across views within a sequence, then averaged with equal weight across sequences. All metrics use the rendered outputs directly, with sampling gaps left intact.

The tri-modal conditioning comparison uses the same trained model and paired test videos. During inference, we set the conditioning latent for appearance, depth or layout to zero in turn, and compare these configurations with those retaining all three conditions or setting all three to zero. Model parameters, text and all other inputs remain fixed. PSNR, SSIM and LPIPS are first aggregated within each camera-specific video clip, then averaged with equal weight across clips.

Generated samples for 3D object detection training follow other vehicles' trajectories in the training-set roadside recordings, with 3D labels transformed from roadside annotations into the target vehicle's coordinate frame. We add generated samples equivalent to $10\%$, $20\%$, $50\%$, $60\%$ and $100\%$ of the original real training frame count, using real-data-only training as the baseline. Both real and generated samples use the six camera views other than the front wide-angle view. For each configuration, we train BEVFormer~\cite{li2022bevformer}, StreamPETR~\cite{wang2023streampetr} and CAPE~\cite{xiong2023cape}, and evaluate AP for cars, buses, pedestrians, motorcycles, bicycles and barriers, together with the six-class mAP, on the same real onboard test set.

\endgroup

\FloatBarrier
\bibliographystyle{unsrtnat}
\begingroup
\small
\bibliography{references}
\endgroup

\end{document}